\documentclass[10pt,twocolumn,letterpaper]{article}

\usepackage{cvpr} 

\usepackage{pifont}
\usepackage{booktabs}
\usepackage{capt-of}
\usepackage{graphicx}
\usepackage[accsupp]{axessibility}
\newcommand{\cmark}{\ding{51}} 
\newcommand{\xmark}{\ding{55}} 

\definecolor{cvprblue}{rgb}{0.21,0.49,0.74}
\usepackage[breaklinks,colorlinks,allcolors=cvprblue]{hyperref}

\def\confName{CVPR}
\def\confYear{2026}

\title{Mind the Hitch: Dynamic Calibration and Articulated Perception for Autonomous Trucks}

\author{
Morui Zhu\textsuperscript{1},
Yongqi Zhu\textsuperscript{1},
Song Fu\textsuperscript{1},
Qing Yang\textsuperscript{1}\\[4pt]
\textsuperscript{1}University of North Texas
}
\begin{document}

\twocolumn[{%
\maketitle
\vspace{-2mm}
\begin{center}
  \begin{minipage}{\textwidth}
    \centering
    \includegraphics[width=\textwidth]{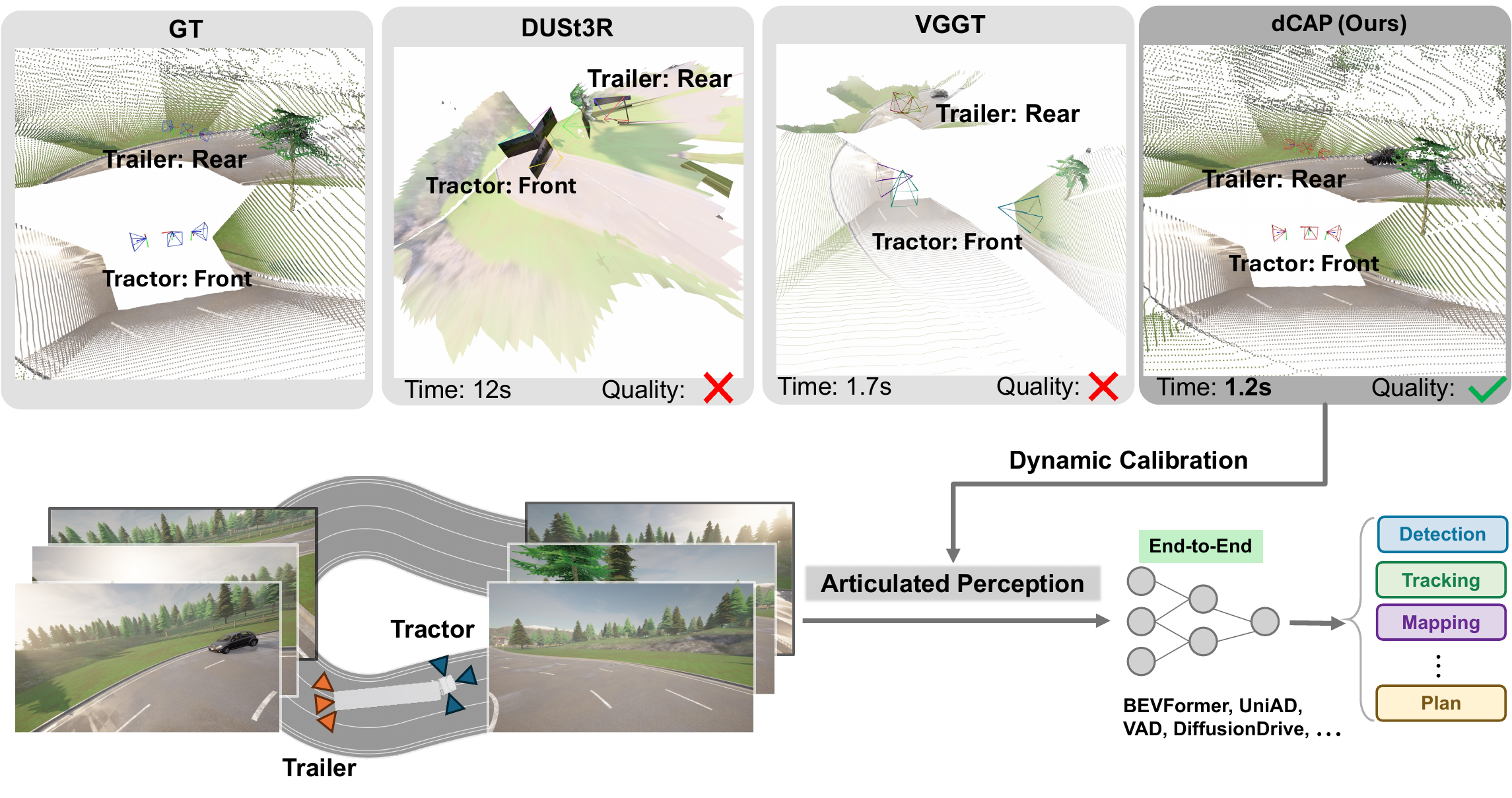}
    \captionof{figure}{Overview of dCAP. We perform online 6-DoF articulated pose estimation for tractor--trailer systems and enable articulated-aware perception. Unlike traditional SfM methods (e.g., COLMAP), dCAP succeeds without requiring a valid static initialization pair.}
    \label{fig:cover_paper}
  \end{minipage}
\end{center}
\vspace{-2mm}
}]

\begin{abstract}

Autonomous trucking poses unique challenges due to articulated tractor–trailer geometry, and time-varying sensor poses caused by the fifth-wheel joint and trailer flex. Existing perception and calibration methods assume static baselines or rely on high-parallax and texture-rich scenes, limiting their reliability under real-world settings. We propose dCAP (dynamic Calibration and Articulated Perception), a vision-based framework that continuously estimates the 6-DoF (degree of freedom) relative pose between tractor and trailer cameras. dCAP employs a transformer with cross-view and temporal attention to robustly aggregate spatial cues while maintaining temporal consistency, enabling accurate perception under rapid articulation and occlusion. Integrated with BEVFormer, dCAP improves 3D object detection by replacing static calibration with dynamically predicted extrinsics. To facilitate evaluation, we introduce STT4AT, a CARLA-based benchmark simulating semi-trailer trucks with synchronized multi-sensor suites and time-varying inter-rig geometry across diverse environments. Experiments demonstrate that dCAP achieves stable, accurate perception while addressing the limitations of static calibration in autonomous trucking. The dataset, development kit, and source code will be publicly released.
\end{abstract}

\section{Introduction}
\label{sec:intro}

Autonomous trucking promises substantial gains in freight safety and efficiency, yet it differs fundamentally from single rigid vehicles in both geometry and operation. 
Long-haul tractors tow articulated trailers whose length, mass, and hinge-based kinematics often introduce off-tracking, rear swing, large turning radii. 
The articulation joint causes the tractor and trailer to constantly move, relative to each other, which changes the positions of sensors mounted on each. 
%
Beyond the geometric challenges, the operational model of freight transport further complicates the problem. 
Tractors and trailers are often owned or maintained by different companies, and a single tractor may attach to multiple trailers during operation.
Therefore, automatic and reliable calibration between the tractor and trailer sensors becomes essential for autonomous truck's perception and control.


\subsection{Research Problems}
Articulated trucks introduce unique challenges for perception and calibration because the tractor and trailer are connected by a moving joint rather than a rigid frame. 
The fifth-wheel coupling creates a time-varying 3D transformation between sensors mounted on the tractor and those on the trailer. 
In real-world settings, this relationship constantly changes due to suspension movement, trailer flex, varying loads, and pitch shifts during braking or on slopes. 
As a result, a calibration that is correct at one moment can become inaccurate just milliseconds later.

These dynamic effects break the fixed-baseline assumption used by most multi-view perception systems~\cite{li2022bevformer, hu2023planning_uniad, jiang2023vad, liao2025diffusiondrive, song2025momad, Zhang_2025_CVPR_bridgeAD}. 
When the articulation angle changes, epipolar geometry drifts, and camera calibration becomes dependent on both the scene and the driving maneuver. 
Small errors in timing or rolling-shutter readout can also lead to large geometric distortions. 
This makes simple fusion of tractor and trailer camera views unreliable, often causing unstable perception and pose estimation. 
Therefore, autonomous trucking requires continuous online estimation of the trailer’s pose relative to the tractor, robust to fast articulation and low-texture scenes, rather than relying on static or occasional offline calibrations.

\subsection{Limitations of Prior Work}
Existing methods for articulated truck perception either rely on strong assumptions or struggle under challenging real-world conditions.
TruckV2X~\cite{xie2025truckv2x} offers a framework for cooperative perception between a tractor and trailer using V2X (Vehicle-to-Everything) communication links. However, it assumes oracle relative poses between the tractor and trailer, which is unrealistic in practice.
Geometry-first approaches~\cite{bundle, locher2016progressive_PMVS, schoenberger2016sfm}, e.g., COLMAP~\cite{schoenberger2016sfm}, can in principle solve the calibration problem by estimating both camera poses and 3D scene structure from multiple images. 
In practice, they struggle under weak parallax, repetitive textures, rolling-shutter effects, and self-occlusion, resulting in inconsistent scale and unstable pose graphs. 
Learning-based geometric methods~\cite{sun2021neuralrecon,lu2025matrix3d, wang2024vggsfm, wang2024dust3r, wang2025vggt}, e.g., VGGT~\cite{wang2025vggt} and DUSt3R~\cite{wang2024dust3r}, improve robustness to photometric variation, but they still fail in rapid articulation, near-field clutter, and texture-poor highway scenarios (Figure~\ref{fig:cover_paper}).

A key limitation of existing methods is that they ignore the rigid structure of tractor–trailer rigs: cameras on each vehicle form fixed rigs, with only the inter-rig transform changing over time. 
We exploit this by predicting the rear trailer camera pose relative to the tractor at each timestep. 
We introduce an end-to-end transformer that directly regresses the dynamic inter-rig pose, enabling online, accurate trailer camera prediction even under challenging articulated maneuvers.

\subsection{Proposed Solution}
We introduce dCAP (\underline{d}ynamic \underline{C}alibration and \underline{A}rticulated \underline{P}erception), a vision-based framework that continuously estimates the relative translation and rotation between tractor and trailer cameras, enabling accurate 3D object detection.

\textbf{Dynamic Calibration.} dCAP employs a transformer-based architecture that first encodes six surrounding RGB views using a Visual Geometry Grounded Transformer (VGGT)~\cite{wang2025vggt} backbone to extract camera-specific tokens capturing spatial geometry. 
A learnable rear-camera query then aggregates cross-view information through a Camera Cross-Attention (CCA) module, attending to the most relevant spatial cues for the trailer region. 
To ensure temporal consistency, a Camera Temporal Self-Attention (CTA) mechanism aligns historical tokens using ego-motion estimates, stabilizing features under articulated motion. 
The aggregated representation is then refined by an adaptive modulation trunk, where pose-dependent normalization dynamically adjusts intermediate features across multiple refinement steps. 
Finally, a lightweight MLP (Multi-Layer Perceptron) head iteratively regresses the 6-DoF (Degrees of Freedom) rear-camera pose in quaternion form, allowing stable and accurate pose estimation even under occlusions and complex articulated dynamics.

\textbf{Articulated Perception.} To assess the impact of dynamic calibration on autonomous driving, we integrate dCAP with BEVFormer \cite{li2022bevformer}, a representative BEV-based detection framework. 
BEVFormer encodes multi-view image features into a unified bird’s-eye-view (BEV) representation and applies a Deformable DETR head \cite{zhu2020deformable} for 3D object detection. 
During inference, we replace the static camera parameters with dCAP’s predicted extrinsics, allowing us to measure how accurate calibration affects detection performance under articulated motion.

\textbf{STT4AT.} To support systematic evaluation, we build a new benchmark in CARLA, called STT4AT (\underline{S}emi-\underline{T}railer \underline{T}ruck \underline{f}or \underline{A}utonomous \underline{T}rucking). 
In this benchmark, we simulate a semi-truck platform that models both the tractor and trailer, each equipped with synchronized multi-sensor suites and capable of recording time-varying inter-rig geometry.
The setup includes six surround-view cameras, a spinning LiDAR, and dual GNSS–IMU units. 
%
We collect data across eight CARLA towns, covering a broad range of driving environments such as highways, urban grids, logistics yards, and terminals. 
Scenarios are designed to induce large articulation angles through challenging maneuvers like U-turns, roundabouts, multi-turn sequences, lane changes, and intersection traversals. 
All sequences follow the nuScenes \cite{caesar2020nuscenes} format, including calibrated intrinsics/extrinsics, 3D bounding boxes for dynamic agents, and high-level semantic maps, ensuring compatibility with existing benchmarks and facilitating cross-task evaluation.

\section{Related Work}
\label{sec:formatting}


\textbf{Semi-trailer Truck Datasets.}
Despite progress in large-scale autonomous driving benchmarks such as KITTI~\cite{Geiger2013IJRR_KITTI}, nuScenes~\cite{caesar2020nuscenes}, Waymo Open~\cite{Sun_2020_CVPR_Waymo}, and Argoverse~\cite{Argoverse2}, few datasets focus on heavy-duty trucks. 
Collecting ground truth for commercial vehicles is challenging due to their length, articulated kinematics, and the need for multi-body calibration, making data costly and less standardized. %
Most truck datasets remain proprietary, limiting public research compared to passenger vehicles.

The MAN TruckScenes dataset~\cite{truckscenes2024} is the first public benchmark for trucks but models the truck–trailer system as a rigid body, ignoring articulation and calibration drift. 
TruckV2X~\cite{xie2025truckv2x} introduces cooperative perception across tractor and trailer, but relies on simulator-provided relative poses and assumes trailer-side computation, making it impractical for real-world use. 
In contrast, the proposed STT4AT provides a public dataset with dynamic articulation and time-varying extrinsics, supporting realistic evaluation of articulated perception, calibration, and planning.

\textbf{Dynamic Calibration.}
Most existing calibration methods target rigid sensor rigs, assuming fixed inter-sensor geometry and operating offline~\cite{8057987_static_2, 8814231_static_3, eker2022real_static_4,herau2024soac,cocheteux2024muli, luu2025rc_AutoCalib,10.1145/3664647.3680572_calibrbev}. 
Examples include UniCal~\cite{yang2024unical_static_0}, which learns differentiable calibration across modalities, and CaLiV~\cite{tahiraj2025calivlidartovehiclecalibrationarbitrary_static_1}, which performs LiDAR-to-vehicle calibration under non-overlapping views. These approaches, however, are unsuitable for articulated systems, where relative poses vary continuously.

Tractor–trailer configurations require dynamic calibration, estimating inter-rig transformations online as the articulation joint moves. 
DSVT~\cite{dong2024dsvt_dynamic_0} addresses this by estimating relative poses via epipolar geometry, but it is limited by stereo constraints and fails under low-texture or large-articulation scenarios. 
Other learning-based methods, such as UDSV~\cite{UDSV} and cascaded visual alignment frameworks~\cite{Unsupervised_deep_image_stitching}, focus on image stitching rather than geometric calibration. 
Thus, dynamic calibration for semi-trailer trucks from raw visual inputs remains largely unsolved.

\section{Benchmark}
\subsection{Semi-trailer Truck Dataset Construction}

To investigate articulated perception and dynamic calibration under realistic tractor–trailer motion, we first reconstructed a dedicated semi-trailer truck dataset in the CARLA~0.9.16 simulator~\cite{Dosovitskiy17}. It consists of 87 scenes from 8 towns and provides multimodal data based on nuScenes~\cite{caesar2020nuscenes} format.
The truck models are adapted from the improved tractor–trailer system~\cite{attard2024autonomousnavigationtractortrailervehicles}, which enable physically consistent articulation and wheel dynamics. 

\textbf{Sensor Setup.}
Each truck is equipped with a synchronized multi-sensor suite composed of six RGB cameras, one LiDAR, and two integrated GPS–IMU module mounted on tractor and trailer, respectively, as summarized in Table~\ref{tab:sensor_data_config}. 
Three cameras were mounted on the tractor’s cabin (front, front-left, and front-right), while three additional cameras were installed on the trailer’s rear (rear, rear-left, and rear-right).
The camera intrinsics and extrinsics on truck remain fixed, whereas the trailer-mounted cameras exhibit continuously varying extrinsics due to hitch rotation. 
%

\begin{table}[t]
\centering
\small
\setlength{\tabcolsep}{6pt}
\begin{tabular}{ll}
\toprule
Sensor & Details \\
\midrule
6x Camera & RGB, 1600 $\times$ 900 resolution, 110$^{\circ}$ FOV \\
1x LiDAR & 128 channels, 3.5M points per second, \\ & 200 m capturing range, -20$^{\circ}$ to 20$^{\circ}$ vertical \\ &  FOV, $\pm$2 cm error \\
2x GPS \& 2x IMU & 20 mm positional error, 2$^{\circ}$ heading error \\
\bottomrule
\end{tabular}
\caption{Sensor specifications.}
\label{tab:sensor_data_config}
\end{table}

\begin{figure*}[!htbp]
    \centering
    \includegraphics[width=1\linewidth]{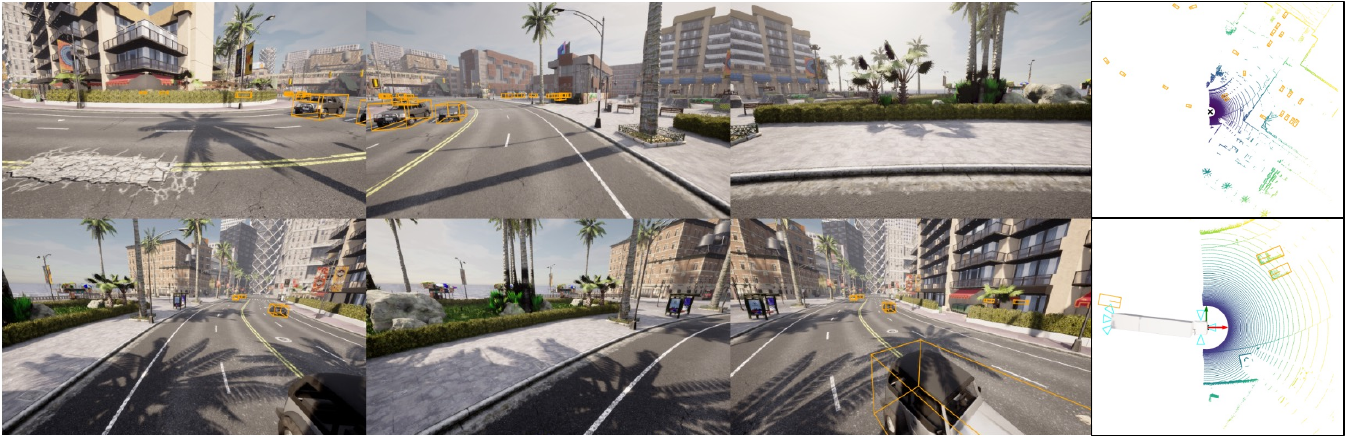}
    \caption{Example from the STT4AT dataset showing six synchronized camera views, the LiDAR point cloud with annotated agents, a BEV illustration of trailer articulation.}
    \label{fig:Examples of the dataset}
\end{figure*}

\textbf{Data Annotation.}
All dynamic agents/objects are annotated with 3D bounding boxes defined by their geometric center $(x, y, z)$, size $(w, l, h)$, and orientation, represented as quaternions $(\hat{w}, \hat{x}, \hat{y}, \hat{z})$. 
Each object maintains a consistent identity across frames to support multi-object tracking~\cite{qin2024towards} and motion forecasting~\cite{zhou2025modeseq}. 
The dataset includes high-resolution semantic maps with multiple layers, showing drivable areas, lane markings, road dividers, sidewalks, and pedestrian crossings. 
In addition to object-level labels, ego-vehicle trajectories and articulated trailer poses are recorded at 10~Hz, ensuring temporal consistency for perception, planning, and dynamic calibration studies.
Figure~\ref{fig:Examples of the dataset} shows a representative example, showing the six synchronized camera views, the LiDAR point cloud with detected 3D bounding boxes, and a BEV illustration of trailer articulation..

\textbf{Scenario Coverage.}
To capture the full spectrum of articulated motion, we collected data from eight CARLA towns (Town01–07 and Town10), covering a wide range of geometric and kinematic configurations. 
Emphasis was placed on scenarios that induce large articulation angles between the tractor and trailer, including U-turns, intersections, roundabouts, lane changes, and multi-turn sequences.

\begin{figure}
    \centering
    \includegraphics[width=1\linewidth]{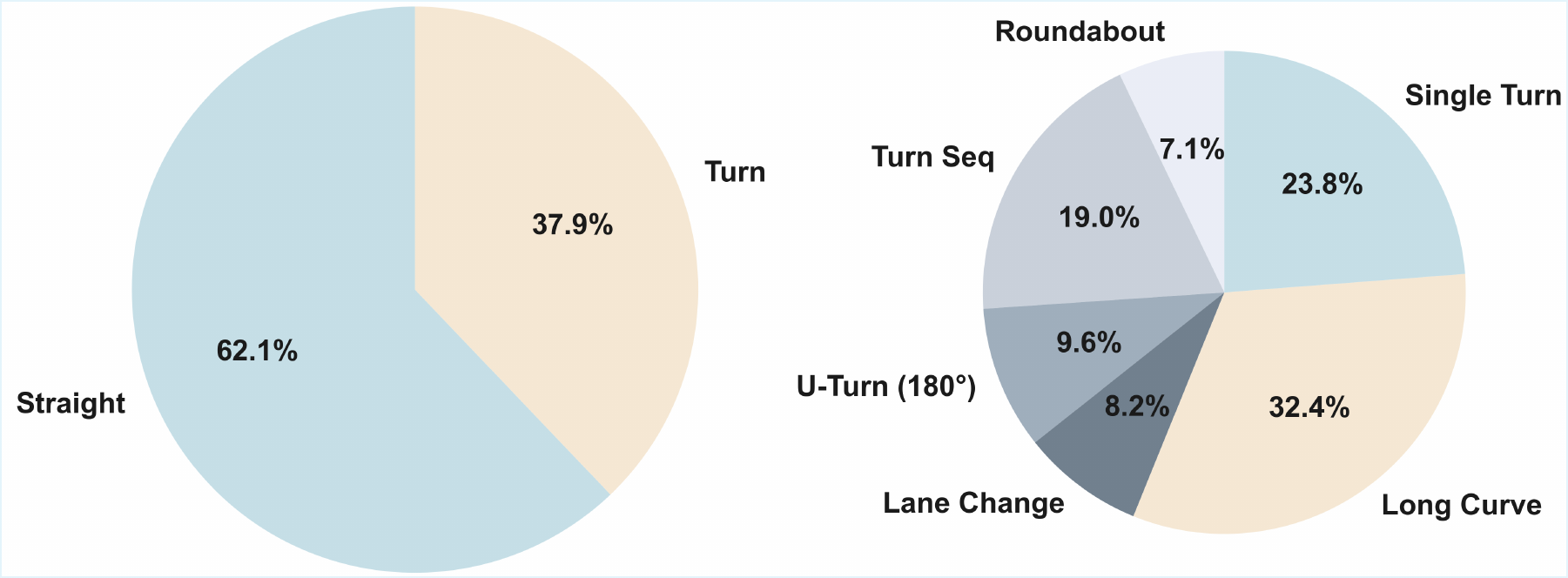}
    \caption{Distribution of annotated frames in the STT4AT dataset. The left chart separates straight and turning maneuvers, while the right details the composition within turning scenarios.}
    \label{fig:dataset_stat}
\end{figure}

%

Among 4,533 annotated frames, turning maneuvers account for 37.9\% of the total. Within this subset, the largest proportions correspond to \textit{Long Curve} (32.4\%) and \textit{Single Turn} (23.8\%), followed by \textit{Turn Sequence} (19.0\%), \textit{U-Turn (180°)} (9.6\%), \textit{Lane Change} (8.2\%), and \textit{Roundabout} (7.1\%). These distributions demonstrate a balanced coverage of both gentle and sharp trailer rotations across diverse geometric layouts.
This comprehensive dataset provides a controlled yet realistic benchmark for evaluating calibration, perception, and planning algorithms under dynamic trailer articulation.

\subsection{Architecture of dCAP}

\begin{figure*}
    \centering
    \includegraphics[width=1\linewidth]{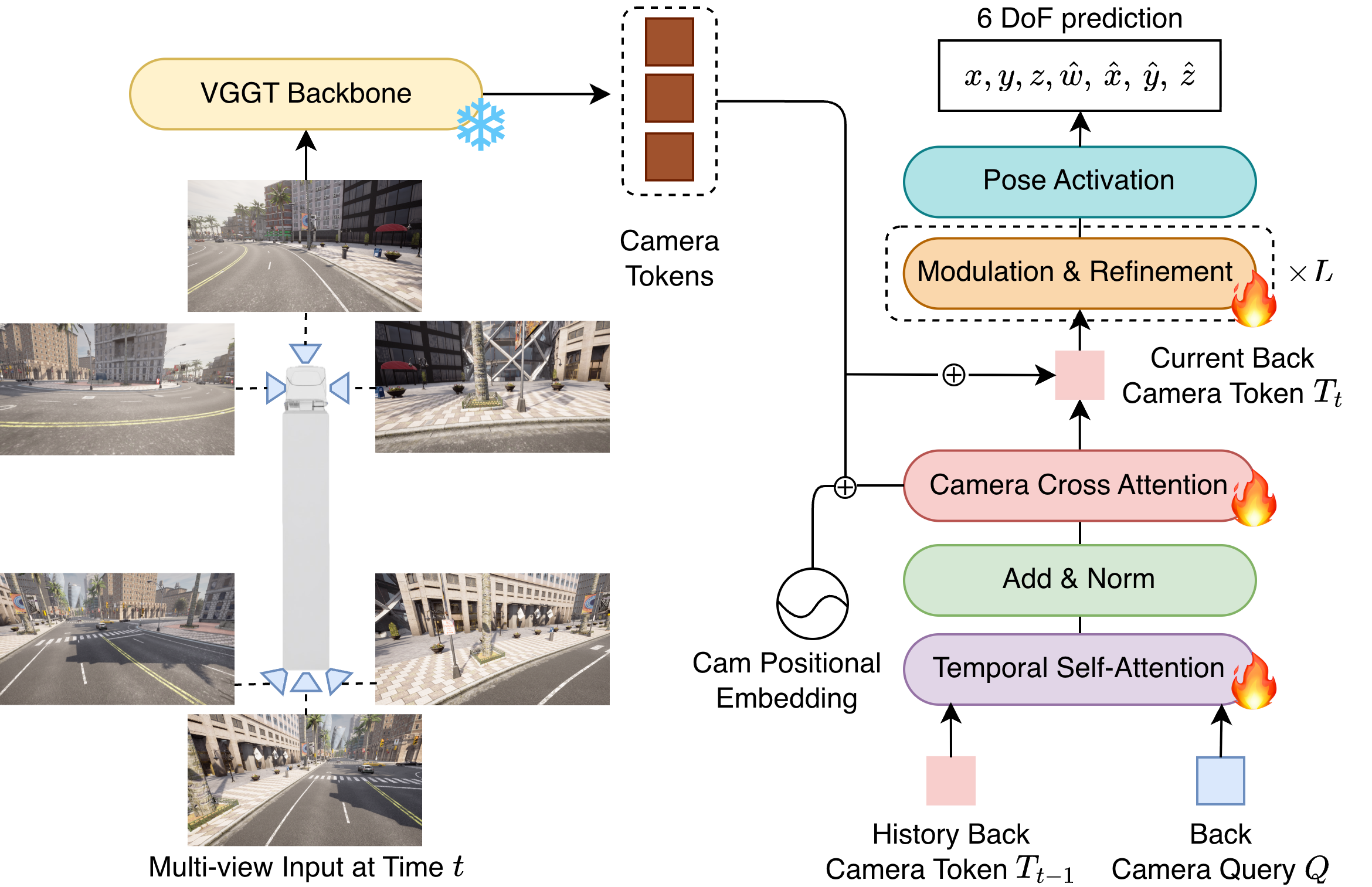}
    \caption{Overview of the proposed architecture. Multi-view images at time $t$ are encoded by a frozen VGGT backbone into camera tokens, while the trainable decoder comprises (a) Camera Temporal Self-Attention (CTA) for fusing the historical token $T_{t-1}$ with the current query $Q$, (b) Camera Cross-Attention (CCA) for attending $Q$ to encoder tokens $\{T_i\}_{i=1}^{6}$, (c) an AdaLN-modulated refinement stack with residual \texttt{Add\&Norm} applied $L$ times.
    }
    \label{fig:pipeline}
\end{figure*}

The dCAP framework aims to predict the articulated trailer rear camera pose at any time given multi-view ego-truck images. 
As illustrated in Figure~\ref{fig:pipeline}, it comprises three major components: (1) a frozen VGGT backbone that encodes synchronized multi-view images into unified geometric tokens, (2) a lightweight decoder with \textit{Camera Cross-Attention} (CCA) to aggregate spatial cues across camera views and \textit{Camera Temporal Self-Attention} (CTA) to ensure temporal coherence under articulation, 
and (3) a direct pose regression head that predicts the trailer’s 6-DoF transformation without explicit geometric optimization. 

\textbf{Multi-view Encoding.}
At each timestep $t$, we obtain six synchronized RGB images surrounding the ego truck. 
All images are fed into VGGT backbone~\cite{wang2025vggt}, producing a set of camera-specific latent features. 
A learnable token is appended to each camera stream, resulting in six camera tokens $\{T_1, T_2, ..., T_6\}$, representing spatially contextualized embeddings of the surrounding scene geometry.

\textbf{Camera Cross Attention.}
To infer the articulated trailer camera pose, we introduce a learnable \textit{rear camera query} $Q$ that interacts with the six encoded camera tokens via a multi-head cross-attention module:
\[
Q' = \mathrm{MHA}(Q, \{T_i\}_{i=1}^{6}, \{T_i\}_{i=1}^{6}),
\]
where $\mathrm{MHA}(\cdot)$ denotes the standard multi-head attention. 
It enables the query to aggregate information across all viewpoints, while attending to the most relevant spatial cues for the trailer region. 
Positional embeddings corresponding to camera indices are added before attention to preserve spatial consistency. 
The cross-attended token $Q'$ is further combined with the rear camera token $T_t$ via a residual connection, preserving the intrinsic rear camera representation, while enriching it with cross-view spatially attended information.

\textbf{Camera Temporal Self-Attention.}
To maintain temporal coherence between consecutive frames, we align the historical rear camera tokens, based on tractor's motion. 
Given the ego poses at times $t{-}1$ and $t$, we compute the incremental motion $\Delta p_t = (\Delta x, \Delta y, \Delta \psi)$, 
where $\Delta \psi$ denotes the yaw change. 
This displacement is projected into the feature space by a linear transformation $\phi_\Delta(\cdot)$:
\[
\tilde{T}_{t-1} = T_{t-1} + \phi_\Delta(\Delta p_t),
\quad 
\phi_\Delta(\Delta p_t) = W_\Delta \Delta p_t + b_\Delta,
\]
where $W_\Delta \in \mathbb{R}^{3 \times d}$ and $b_\Delta$ are learnable parameters. 
This operation ensures that past tokens are geometrically aligned with the current ego coordinate frame before temporal fusion. 
Empirically, such pose-aware alignment significantly stabilizes the historical context and prevents feature drift under sharp turns or articulated motion.
After alignment, the current global token $G_t$ interacts with the aligned historical representation $\tilde{T}_{t-1}$ through a multi-head temporal self-attention layer, 
which propagates temporal context and smooths frame-to-frame predictions:
\[
G'_t = G_t + \mathrm{MHA}(G_t, \tilde{T}_{t-1}, \tilde{T}_{t-1}).
\]
 This temporal interaction encourages continuity in both spatial reasoning and pose estimation, allowing the model to exploit motion cues from recent frames while remaining robust to partial occlusions.

\textbf{Modulation and Refinement.}
The aggregated representation is then processed by a modulation--refinement head with $L$ stacked transformer blocks. 
Each block applies adaptive layer normalization~\cite{peebles2024dit} followed by learned affine modulation and a gating residual:
\[
\hat{x} = \gamma \odot \big(\mathrm{AdaLN}(x) \odot (1+\beta) + \alpha \big) + x,
\]
where $(\alpha, \beta, \gamma) \in \mathbb{R}^{d}$ are per-channel shift, scale, and gate parameters predicted from the current pose embedding. 
This design adapts intermediate features to the evolving pose estimate and stabilizes multi-step refinement.

\section{Experiments}

\subsection{Training Details}
All sequences in STT4AT are randomly split into training and validation sets with an 8:2 ratio at the sequence level.
Both training and inference are performed on a single NVIDIA RTX~A6000 GPU. 
During training, the encoder remains frozen, while only the decoder components are optimized, including the CTA, CCA, and modulation–refinement head. 
The model is trained for 24~epochs using the Adam optimizer with an initial learning rate of $1\times10^{-4}$ and a batch size of~4. 
In the CTA module, the temporal queue length is set to~3 to capture motion information across consecutive frames. 
The refinement module performs 3 iterative refinement steps. 
The overall objective combines translation and rotation losses with equal weighting:
\[
L = w_\text{trans} L_\text{trans} + w_\text{rot} L_\text{rot},
\]
where both $L_\text{trans}$ and $L_\text{rot}$ are computed using the $\ell_1$ formulation, and $w_\text{trans}{=}w_\text{rot}{=}1.0$.

\subsection{Metrics}

For trailer pose estimation, all results are reported in metric scale. 
We measure the overall translation error $\Delta_T$ and its axis-wise components $(\Delta_x, \Delta_y, \Delta_z)$. 
Orientation is evaluated using the RRA (Relative Rotation Accuracy), which computes the mean rotational deviation between the predicted and ground-truth rotation matrices $\hat{R}_t$ and $R_t$ as follow.
\[
\text{RRA} = \arccos\!\left(\tfrac{1}{2}\,\mathrm{tr}\!\left(\hat{R}_t^{\top}R_t\right)-1\right).
\]
Since all three trailer rear cameras move as a rigid rig, we estimate the pose of a single rear camera and derive the other two (rear-left and rear-right) poses using the known intra-trailer transformations.

For perception tasks like 3D object detection, we adopt BEVFormer~\cite{li2022bevformer} as the baseline detector and feed it with dynamically calibrated camera extrinsics during inference.
Evaluation follows the nuScenes~\cite{caesar2020nuscenes} protocol, reporting mean Average Precision (mAP$\uparrow$), normalized detection score (NDS$\uparrow$), and individual error metrics including Average Translation Error (ATE$\downarrow$), Average Scale Error (ASE$\downarrow$), Average Orientation Error (AOE$\downarrow$), Average Velocity Error (AVE$\downarrow$), and Average Attribute Error (AAE$\downarrow$). 
We also provide AP at different spatial thresholds (AP@0.5m, AP@1.0m, AP@2.0m, AP@4.0m) to quantify the spatial precision under varying articulation magnitudes.


\subsection{Baseline}
To provide a comprehensive comparison, we include several baseline configurations to represent different calibration strategies.
Static calibration assumes a fixed tractor–trailer geometry and relies on a single-shot extrinsic calibration.
This setting ignores articulation and therefore serves as a lower bound in our evaluation.
A GNSS-IMU Kalman Filter (KF) is used as a classical sensor-fusion baseline under realistic noise (20\,mm position, $2^\circ$ heading).
For VGGT~\cite{wang2025vggt}, DUSt3R~\cite{wang2024dust3r}, and COLMAP~\cite{schoenberger2016sfm}, direct trailer-pose prediction is not supported, as all methods produce a normalized scale.  
To enable comparison, we first estimate the relative transform between the front tractor camera and the rear trailer camera from their reconstructed poses.
Then, we scale the result using a factor derived from the ground-truth scale.  
The resulting transform is converted into the metric trailer-to-tractor calibration used by BEVFormer.  
This procedure allows us to evaluate how well geometry-based methods generalize to articulated systems when their outputs are adapted to metric scale.

\subsection{Trailer Pose Estimation}  
As shown in Table~\ref{tab:pose_pred_quan}, dCAP consistently outperforms all baselines, including geometry-based methods (VGGT~\cite{wang2025vggt}, DUSt3R~\cite{wang2024dust3r}, COLMAP~\cite{schoenberger2016sfm}) and the GNSS-IMU Kalman Filter.
Geometry-based approaches degrade under articulation and limited parallax, while the Kalman Filter accumulates error due to the absence of explicit articulation modeling.
Static calibration yields large translation errors due to its rigid-body assumption.
In contrast, all variants of dCAP reduce both translation and rotation errors.

The dCAP with CCA module achieves the low rotational error ($RRA{=}0.048$), indicating its strong ability to integrate complementary spatial cues across multiple camera views. By attending to encoder tokens $\{T_i\}_{i=1}^{6}$, CCA aggregates cross-view geometric evidence that constrains the trailer orientation even under asymmetric viewpoints and partial occlusions. This spatial reasoning stabilizes angular estimation and reduces the drift typically observed when the rear trailer region is only partially visible.

The dCAP with CTA module attains the smallest translation error ($\Delta_T{=}0.452$), demonstrating its advantage in temporal consistency. By aligning the previous back-camera token $T_{t-1}$ with the current ego pose and performing temporal self-attention, CTA effectively propagates motion cues and smooths frame-to-frame variations. Such pose-aware temporal fusion is particularly beneficial when the trailer undergoes rapid articulation or transient occlusion, ensuring coherent translation estimation across consecutive frames.

Combining both modules yields achieves the lowest rotation error ($RRA{=}0.042$) and improved axis-wise accuracy (e.g., $\Delta_x{=}0.061$), while maintaining the minimal translation error ($\Delta_T{=}0.452$).
This shows that spatial and temporal modeling contribute complementary benefits to articulated pose estimation.

\begin{table}[t]
\centering
\small
\resizebox{\columnwidth}{!}{
\setlength{\tabcolsep}{6pt}
\begin{tabular}{crrrrrr}
\toprule
Method & $\Delta_T\downarrow$ & $\Delta_x\downarrow$ & $\Delta_y\downarrow$ & $\Delta_z\downarrow$ & RRA$\downarrow$ \\
\midrule
Static Calibration & 1.284 & 0.210 & 1.120 & 0.356 & 0.148 \\
COLMAP~\cite{schoenberger2016sfm}$^{\dagger}$ & - & - & - & - & - \\
VGGT~\cite{wang2025vggt} & 6.040 & 2.761 & 3.082 & 3.634 & 0.309 \\
DUSt3R~\cite{wang2024dust3r} & 8.625 & 4.664 & 5.080 & 2.953 & 0.578 \\
GNSS-IMU (Kalman Filter) & 1.379 & 0.309 & 1.116 & 0.431 & 0.129 \\
\midrule
dCAP (w/o CCA, w/o CTA) & 0.632 & 0.076 & 0.600 & 0.087 & 0.073 \\
dCAP (w/ CCA, w/o CTA) & \underline{0.505} & \underline{0.069} & 0.475 & \textbf{0.074} & \underline{0.048} \\
dCAP (w/o CCA, w/ CTA) & \textbf{0.452} & 0.125 & \textbf{0.395} & 0.090 & 0.058 \\
dCAP (w/ CCA, w/ CTA) & \textbf{0.452} & \textbf{0.061} & \underline{0.421} & \underline{0.085} & \textbf{0.042} \\
\bottomrule
\end{tabular}
}
\caption{
Quantitative results of trailer camera pose prediction under different methods. Note that $^\dagger$ fails to reconstruct due to the lack of a valid initial image pair.
}
\label{tab:pose_pred_quan}
\end{table}

\subsection{3D Object Detection}
We integrate dCAP into BEVFormer to evaluate object detection performance on autonomous trucks. 
During inference, the predicted trailer rear-camera pose is first converted into metric scale and propagated to the other two trailer-mounted cameras via known intra-rig extrinsics.
These dynamically estimated camera parameters are then fed into BEVFormer through its standard calibration interface to generate BEV features and perform object detection.

Quantitative results are reported in Table~\ref{tab:res_detection} where dCAP clearly outperforms all geometry-based and static baselines.  
Previous methods such as VGGT, DUSt3R, and COLMAP degrade under articulation and limited overlap, while static calibration and tractor-only configurations suffer from rigid-body assumptions that fail to model trailer motion.  
By contrast, dynamically estimated extrinsics enable stable BEV feature alignment, yielding substantial gains in precision and orientation accuracy. 
Among the proposed modules, CCA achieves strong detection accuracy (AP = 0.102) with low orientation error, while CTA maintains temporal consistency with competitive translation metrics.
Combining both modules further improves performance, achieving the highest AP (0.103) and the lowest orientation error while reducing scale and velocity errors.
Although AP remains low, this is expected because BEVFormer is inherently designed for rigid vehicles with fixed extrinsics, whereas truck combines high-mounted, pitched tractor cameras with continuously moving trailer cameras. 
%
%
Overall, dCAP preserves detection accuracy across diverse articulation scenarios, reducing the gap to the ground-truth upper bound.

\begin{table*}[t]
\centering
\small
\resizebox{\textwidth}{!}{
\setlength{\tabcolsep}{6pt}
\begin{tabular}{crrrrrrrrrrrr}
\toprule
Method & AP$\uparrow$ & NDS$\uparrow$ & ATE$\downarrow$ & ASE$\downarrow$ & AOE$\downarrow$ & AVE$\downarrow$ & AAE$\downarrow$ & AP@0.5m$\uparrow$ & AP@1.0m$\uparrow$ & AP@2.0m$\uparrow$ & AP@4.0m$\uparrow$ \\
\midrule
Static Calibration & 0.058 & 0.033 & 0.734 & 0.176 & 0.153 & 2.419 & 0.237 & 0.0000 & 0.0188 & 0.0728 & 0.1417 \\
COLMAP~\cite{schoenberger2016sfm}$^\dagger$ & - & - & - & - & - & - & - & - & - & - & - \\
Tractor only & 0.049 & 0.032 & 0.705 & 0.183 & 0.198 & 2.572 & 0.267 & 0.0000 & 0.0159 & 0.0623 & 0.1173 \\
VGGT~\cite{wang2025vggt} & 0.033 & 0.031 & 0.671 & 0.181 & 0.202 & 2.619 & 0.251 & 0.0000 & 0.0115 & 0.0442 & 0.0783 \\
DUSt3R~\cite{wang2024dust3r} & 0.034 & 0.031 & 0.711 & 0.182 & 0.219 & 2.682 & 0.256 & 0.0000 & 0.0092 & 0.0422 & 0.0839 \\
dCAP (w/o CCA, w/o CTA) & 0.084 & 0.034 & 0.710 & 0.172 & 0.132 & \underline{2.328} & 0.222 & 0.0007 & 0.0386 & 0.1105 & 0.1863 \\ 
dCAP (w/ CCA, w/o CTA) & \underline{0.102} & \textbf{0.036} & \textbf{0.657} & \underline{0.170} & \underline{0.118} & 2.330 & \underline{0.212} & \textbf{0.0072} & \textbf{0.0598} & \underline{0.1391} & \underline{0.2032} \\ 
dCAP (w/o CCA, w/ CTA) & 0.094 & \underline{0.035} & 0.697 & \underline{0.170} & 0.122 & 2.349 & \textbf{0.211} & 0.0012 & 0.0478 & 0.1293 & 0.1962 \\ 
dCAP (w/ CCA, w/ CTA) & \textbf{0.103} & \textbf{0.036} & \underline{0.675} & \textbf{0.169} & \textbf{0.116} & \textbf{2.312} & \underline{0.212} & \underline{0.0047} & \underline{0.0586} & \textbf{0.1444} & \textbf{0.2052} \\ 
\midrule
GT (upper bound) & 0.129 & 0.039 & 0.513 & 0.168 & 0.105 & 2.258 & 0.209 & 0.0349 & 0.1016 & 0.1680 & 0.2125 \\
\bottomrule
\end{tabular}
}
\caption{
Quantitative results of 3D object detection under different attention configurations and baselines. Note that $^\dagger$ fails to reconstruct due to the lack of a valid initial image pair.
}
\label{tab:res_detection}
\end{table*}

\subsection{Ablation Studies}
\textbf{Camera Prediction Under Different Scenarios.}
To systematically investigate how different attention mechanisms contribute to trailer pose estimation across various motion patterns, we analyze their performance under four representative scenarios: Straight, Roundabout, U-turn, and Multi-Turn.
The motivation stems from the complementary characteristics of the two modules: CCA emphasizes spatial alignment and performs robustly in steady, low-articulation scenes, while CTA leverages temporal consistency that enhances both translation and rotation stability under high-articulation motion.

Quantitatively, the results show that in low-articulation scenarios such as Straight and Multi-Turn, the translational advantage of CTA over CCA is minor ($-11.2\%$ in Straight), while CCA even surpasses CTA by $14.6\%$ in Multi-Turn. 
This indicates that both mechanisms perform comparably when articulation angles are small, while CCA exhibits slightly better performance.
In high-articulation scenarios such as U-turn and Roundabout, however, the contribution of temporal reasoning becomes substantially more pronounced.
CTA achieves a $-36.8\%$ and $-29.6\%$ reduction in translation error compared to CCA, confirming that dynamic temporal fusion is crucial when trailer pose changes rapidly.
By contrast, the rotational gap is modest and even slightly favors CCA: in Roundabout and U-turn, CTA’s $RRA$ is higher than CCA by $8.2\%$ and $9.9\%$, i.e., an order of magnitude smaller than CTA’s $29.6\%$–$36.8\%$ translation gains.

Overall, these comparisons reveal a clear pattern of specialization.
CCA offers robust and consistent performance in scenarios characterized by smooth, continuous motion, where geometric correspondence dominates.
CTA, on the other hand, excels in large-angle maneuvers that require temporal smoothing and motion-aware refinement.


\begin{table}[t]
\centering
\small
\setlength{\tabcolsep}{6pt}
\begin{tabular}{ccrrrrr}
\toprule
CCA & CTA & $\Delta_T\downarrow$ & $\Delta_x\downarrow$ & $\Delta_y\downarrow$ & $\Delta_z\downarrow$ & $RRA\downarrow$ \\
\midrule
\xmark & \xmark & 0.926 & \textbf{0.037} & 0.913 & 0.079 & 0.071 \\  
\cmark & \xmark & \underline{0.517} & \underline{0.046} & \underline{0.501} & \textbf{0.051} & \textbf{0.051} \\  
\xmark & \cmark & \textbf{0.459} & 0.105 & \textbf{0.430} & \underline{0.052} & \underline{0.058} \\  
\bottomrule
\end{tabular}
\caption{
Trailer pose estimation under the Straight scenario. 
}
\end{table}

\begin{table}[t]
\centering
\small
\setlength{\tabcolsep}{6pt}
\begin{tabular}{ccrrrrr}
\toprule
CCA & CTA & $\Delta_T\downarrow$ & $\Delta_x\downarrow$ & $\Delta_y\downarrow$ & $\Delta_z\downarrow$ & $RRA\downarrow$ \\
\midrule
\xmark & \xmark & 0.850 & \underline{0.136} & 0.803 & \underline{0.097} & 0.095 \\  
\cmark & \xmark & \underline{0.675} & \textbf{0.119} & \underline{0.634} & \textbf{0.094} & \textbf{0.061} \\  
\xmark & \cmark & \textbf{0.475} & 0.168 & \textbf{0.398} & 0.100 & \underline{0.066} \\  
\bottomrule
\end{tabular}
\caption{
Trailer pose estimation under the Roundabout scenario. 
}
\end{table}


\begin{table}[t]
\centering
\small
\setlength{\tabcolsep}{6pt}
\begin{tabular}{ccccccccc}
\toprule
CCA & CTA & $\Delta_T\downarrow$ & $\Delta_x\downarrow$ & $\Delta_y\downarrow$ & $\Delta_z\downarrow$ & $RRA\downarrow$ \\
\midrule
\xmark & \xmark & 1.212 & \textbf{0.159} & 1.175 & 0.154 & 0.112 \\  
\cmark & \xmark & \underline{1.117} & \underline{0.188} & \underline{1.083} & \underline{0.119} & \textbf{0.091} \\  
\xmark & \cmark & \textbf{0.706} & 0.202 & \textbf{0.642} & \textbf{0.097} & \underline{0.100} \\  
\bottomrule
\end{tabular}
\caption{
Trailer pose estimation under the U-turn scenario. 
}
\end{table}


\begin{table}[t]
\centering
\small
\setlength{\tabcolsep}{6pt}
\begin{tabular}{ccrrrrr}
\toprule
CCA & CTA & $\Delta_T\downarrow$ & $\Delta_x\downarrow$ & $\Delta_y\downarrow$ & $\Delta_z\downarrow$ & $RRA\downarrow$ \\
\midrule
\xmark & \xmark & 0.520 & \underline{0.081} & 0.459 & 0.140 & 0.066 \\  
\cmark & \xmark & \textbf{0.361} & \textbf{0.069} & \textbf{0.286} & \textbf{0.118} & \textbf{0.037} \\  
\xmark & \cmark & \underline{0.423} & 0.140 & \underline{0.325} & \underline{0.129} & \underline{0.058} \\  
\bottomrule
\end{tabular}
\caption{
Trailer pose estimation under the Multi-Turn scenario. 
}
\end{table}


\textbf{3D Object Detection Under Different Scenarios.}
We further examine how different attention mechanisms affect 3D object detection across various articulated driving scenarios.
Quantitatively, the results exhibit a pattern consistent with the pose estimation analysis.

As shown in Table~\ref{tab:straight_ablation}, in low-articulation scenes such as Straight and Multi-Turn, CCA clearly dominates.
In the Straight case, for example, CCA improves mAP from 0.0497 to 0.0549, a gain of 10.5\% over CTA, while reducing ATE from 0.9632 to 0.9561 (a 0.7\% improvement).
Similarly, in Multi-Turn (in Table~\ref{tab:multi_turn_ablation}), CCA outperforms CTA by 4.9\% in mAP (0.0496 vs.\ 0.0473) and achieves lower orientation and velocity errors (AOE $\downarrow$ 0.8985 vs.\ 0.8999; AVE $\downarrow$ 1.1482 vs.\ 1.1593).
These results indicate that spatial cross-view alignment is sufficient to maintain geometric consistency when the articulation angle remains small and motion transitions are smooth.

Conversely, in high-articulation maneuvers such as U-turn and Roundabout, CTA exhibits clear advantages.
For the Roundabout scenario (in Table~\ref{tab:roundabout_ablation}), CTA improves mAP by 22.5\% over CCA (0.0397 vs.\ 0.0324) and achieves lower AOE (0.9291 vs.\ 0.9326, a 0.4\% reduction).
A similar trend is observed in U-turn (in Table~\ref{tab:ablation_detection_uturn}), where CTA improves mAP by 3.9\% and reduces rotational error (AOE $\downarrow$ 0.8981 vs.\ 0.8988) and attribute error (AAE $\downarrow$ 0.8905 vs.\ 0.8939).
These consistent gains demonstrate that temporal modeling is crucial when trailer motion involves abrupt articulation, partial occlusion, or rapid viewpoint change.

Overall, a consistent pattern is observed across both pose estimation and detection experiments.
CCA excels in structured and continuous motion where geometric correspondence dominates, while CTA proves more effective in complex turning sequences requiring temporal smoothing and motion-aware adaptation.


\begin{table}[t]
\centering
\small
\resizebox{\columnwidth}{!}{
\begin{tabular}{ccrrrrrr}
\toprule
CCA & CTA & mAP$\uparrow$ & ATE$\downarrow$ & ASE$\downarrow$ & AOE$\downarrow$ & AVE$\downarrow$ & AAE$\downarrow$ \\
\midrule
\xmark & \xmark & 0.0481 & \underline{0.9598} & \underline{0.9175} & 0.8972 & 1.0135 & 0.9104 \\ 
\cmark & \xmark & \textbf{0.0549} & \textbf{0.9561} & 0.9176 & \textbf{0.8964} & \underline{1.0064} & \textbf{0.9091} \\ 
\xmark & \cmark & \underline{0.0497} & 0.9632 & \textbf{0.9173} & \underline{0.8966} & \textbf{1.0051} & \underline{0.9103} \\ 
\midrule
\multicolumn{2}{c}{GT (upper bound)} & 0.0686 & 0.9414 & 0.9174 & 0.8960 & 1.0035 & 0.9096 \\ 
\bottomrule
\end{tabular}
}
\caption{
Detection performance under the Straight scenario. 
}
\label{tab:straight_ablation}
\end{table}


\begin{table}[t]
\centering
\small
\resizebox{\columnwidth}{!}{
\begin{tabular}{ccrrrrrr}
\toprule
CCA & CTA & mAP$\uparrow$ & ATE$\downarrow$ & ASE$\downarrow$ & AOE$\downarrow$ & AVE$\downarrow$ & AAE$\downarrow$ \\
\midrule
\xmark & \xmark & 0.0290 & 0.9800 & \underline{0.9177} & 0.9402 & \textbf{1.1919} & 0.9255 \\ 
\cmark & \xmark & \underline{0.0324} & \underline{0.9770} & \textbf{0.9175} & \underline{0.9326} & \underline{1.1998} & \underline{0.9206} \\ 
\xmark & \cmark & \textbf{0.0397} & \textbf{0.9730} & 0.9182 & \textbf{0.9291} & 1.2058 & \textbf{0.9152} \\ 
\midrule
\multicolumn{2}{c}{GT (upper bound)} & 0.0507 & 0.9564 & 0.9180 & 0.9229 & 1.1704 & 0.9203 \\ 
\bottomrule
\end{tabular}
}
\caption{
Detection performance under the Roundabout scenario. 
}
\label{tab:roundabout_ablation}
\end{table}

\begin{table}[t]
\centering
\small
\resizebox{\columnwidth}{!}{
\begin{tabular}{ccrrrrrr}
\toprule
CCA & CTA & mAP$\uparrow$ & ATE$\downarrow$ & ASE$\downarrow$ & AOE$\downarrow$ & AVE$\downarrow$ & AAE$\downarrow$ \\
\midrule
\xmark & \xmark & 0.0416 & 0.9760 & 0.9183 & 0.8996 & 1.1722 & \underline{0.8925} \\ 
\cmark & \xmark & \underline{0.0464} & \underline{0.9689} & \textbf{0.9180} & \underline{0.8988} & \textbf{1.1518} & 0.8939 \\ 
\xmark & \cmark & \textbf{0.0482} & \textbf{0.9683} & \underline{0.9179} & \textbf{0.8981} & \underline{1.1534} & \textbf{0.8905} \\ 
\midrule
\multicolumn{2}{c}{GT (upper bound)} & 0.0663 & 0.9424 & 0.9176 & 0.8984 & 1.1604 & 0.8901 \\ 
\bottomrule
\end{tabular}
}
\caption{
Detection performance under the U-turn scenario. 
}
\label{tab:ablation_detection_uturn}
\end{table}

\begin{table}[t]
\centering
\small
\resizebox{\columnwidth}{!}{
\begin{tabular}{ccrrrrrr}
\toprule
CCA & CTA & mAP$\uparrow$ & ATE$\downarrow$ & ASE$\downarrow$ & AOE$\downarrow$ & AVE$\downarrow$ & AAE$\downarrow$ \\
\midrule
\xmark & \xmark & 0.0437 & 0.9641 & \textbf{0.9166} & \underline{0.8988} & \textbf{1.1474} & \underline{0.8770} \\ 
\cmark & \xmark & \textbf{0.0496} & \underline{0.9627} & \textbf{0.9166} & \textbf{0.8985} & \underline{1.1482} & \textbf{0.8753} \\ 
\xmark & \cmark & \underline{0.0473} & \textbf{0.9606} & \underline{0.9168} & 0.8999 & 1.1593 & 0.8783 \\ 
\midrule
\multicolumn{2}{c}{GT (upper bound)} & 0.0553 & 0.9509 & 0.9163 & 0.8985 & 1.1435 & 0.8758 \\ 
\bottomrule
\end{tabular}
}
\caption{
Detection performance under the Multi-Turn scenario. 
}
\label{tab:multi_turn_ablation}
\end{table}

\section{Conclusions}

We present a unified benchmark including a semi-trailer truck dataset STT4AT and a vision-based end-to-end framework dCAP that performs dynamic calibration and articulated perception for tractor–trailer systems.  
It achieves state-of-the-art results on dynamic calibration and downstream articulated perception directly from multi-view images.
Its simplicity, efficiency, and robustness under large articulation make it a strong foundation for efficient articulated perception and future research in motion-aware autonomous trucking.

{
    \small
    \bibliographystyle{ieeenat_fullname}
    \bibliography{main}

@String(CVPR= {IEEE Conf. Comput. Vis. Pattern Recog.})

@String(ICCV= {Int. Conf. Comput. Vis.})

@String(ECCV= {Eur. Conf. Comput. Vis.})

@String(CVPR  = {CVPR})

@String(ICCV  = {ICCV})

@String(ECCV  = {ECCV})

@misc{attard2024autonomousnavigationtractortrailervehicles,
      title={Autonomous Navigation of Tractor-Trailer Vehicles through Roundabout Intersections}, 
      author={Daniel Attard and Josef Bajada},
      year={2024},
      eprint={2401.04980},
      archivePrefix={arXiv},
      primaryClass={cs.RO},
      url={https://arxiv.org/abs/2401.04980}, 
}

@inproceedings{caesar2020nuscenes,
  title={nuscenes: A multimodal dataset for autonomous driving},
  author={Caesar, Holger and Bankiti, Varun and Lang, Alex H and Vora, Sourabh and Liong, Venice Erin and Xu, Qiang and Krishnan, Anush and Pan, Yu and Baldan, Giancarlo and Beijbom, Oscar},
  booktitle={Proceedings of the IEEE/CVF conference on computer vision and pattern recognition},
  pages={11621--11631},
  year={2020}
}

@article{Geiger2013IJRR_KITTI,
  author = {Andreas Geiger and Philip Lenz and Christoph Stiller and Raquel Urtasun},
  title = {Vision meets Robotics: The KITTI Dataset},
  journal = {International Journal of Robotics Research (IJRR)},
  year = {2013}
}

@InProceedings{Sun_2020_CVPR_Waymo,
author = {Sun, Pei and Kretzschmar, Henrik and Dotiwalla, Xerxes and Chouard, Aurelien and Patnaik, Vijaysai and Tsui, Paul and Guo, James and Zhou, Yin and Chai, Yuning and Caine, Benjamin and Vasudevan, Vijay and Han, Wei and Ngiam, Jiquan and Zhao, Hang and Timofeev, Aleksei and Ettinger, Scott and Krivokon, Maxim and Gao, Amy and Joshi, Aditya and Zhang, Yu and Shlens, Jonathon and Chen, Zhifeng and Anguelov, Dragomir},
title = {Scalability in Perception for Autonomous Driving: Waymo Open Dataset},
booktitle = {Proceedings of the IEEE/CVF Conference on Computer Vision and Pattern Recognition (CVPR)},
month = {June},
year = {2020}
}

@INPROCEEDINGS {Argoverse2,
  author = {Benjamin Wilson and William Qi and Tanmay Agarwal and John Lambert and Jagjeet Singh and Siddhesh Khandelwal and Bowen Pan and Ratnesh Kumar and Andrew Hartnett and Jhony Kaesemodel Pontes and Deva Ramanan and Peter Carr and James Hays},
  title = {Argoverse 2: Next Generation Datasets for Self-driving Perception and Forecasting},
  booktitle = {Proceedings of the Neural Information Processing Systems Track on Datasets and Benchmarks (NeurIPS Datasets and Benchmarks 2021)},
  year = {2021}
}

@inproceedings{truckscenes2024,
 title = {MAN TruckScenes: A multimodal dataset for autonomous trucking in diverse conditions},
 author = {Fent, Felix and Kuttenreich, Fabian and Ruch, Florian and Rizwin, Farija and Juergens, Stefan and Lechermann, Lorenz and Nissler, Christian and Perl, Andrea and Voll, Ulrich and Yan, Min and Lienkamp, Markus},
 booktitle = {Advances in Neural Information Processing Systems},
 editor = {A. Globerson and L. Mackey and D. Belgrave and A. Fan and U. Paquet and J. Tomczak and C. Zhang},
 pages = {62062--62082},
 publisher = {Curran Associates, Inc.},
 url = {https://proceedings.neurips.cc/paper_files/paper/2024/file/71ac06f0f8450e7d49063c7bfb3257c2-Paper-Datasets_and_Benchmarks_Track.pdf},
 volume = {37},
 year = {2024}
}

@inproceedings{wang2025vggt,
  title={VGGT: Visual Geometry Grounded Transformer},
  author={Wang, Jianyuan and Chen, Minghao and Karaev, Nikita and Vedaldi, Andrea and Rupprecht, Christian and Novotny, David},
  booktitle={Proceedings of the IEEE/CVF Conference on Computer Vision and Pattern Recognition},
  year={2025}
}

@inproceedings{wang2024dust3r,
  title={Dust3r: Geometric 3d vision made easy},
  author={Wang, Shuzhe and Leroy, Vincent and Cabon, Yohann and Chidlovskii, Boris and Revaud, Jerome},
  booktitle={Proceedings of the IEEE/CVF Conference on Computer Vision and Pattern Recognition},
  pages={20697--20709},
  year={2024}
}

@inproceedings{schoenberger2016sfm,
    author={Sch\"{o}nberger, Johannes Lutz and Frahm, Jan-Michael},
    title={Structure-from-Motion Revisited},
    booktitle={Conference on Computer Vision and Pattern Recognition (CVPR)},
    year={2016},
}

@article{xie2025truckv2x,
  title={TruckV2X: A Truck-Centered Perception Dataset},
  author={Xie, Tenghui and Song, Zhiying and Wen, Fuxi and Li, Jun and Liu, Guangzhao and Zhao, Zijian},
  journal={IEEE Robotics and Automation Letters},
  year={2025},
  publisher={IEEE}
}

@inproceedings{Dosovitskiy17,
  title = {{CARLA}: {An} Open Urban Driving Simulator},
  author = {Alexey Dosovitskiy and German Ros and Felipe Codevilla and Antonio Lopez and Vladlen Koltun},
  booktitle = {Proceedings of the 1st Annual Conference on Robot Learning},
  pages = {1--16},
  year = {2017}
}

@inproceedings{
zhu2020deformable,
title={Deformable {\{}DETR{\}}: Deformable Transformers for End-to-End Object Detection},
author={Xizhou Zhu and Weijie Su and Lewei Lu and Bin Li and Xiaogang Wang and Jifeng Dai},
booktitle={International Conference on Learning Representations},
year={2021},
url={https://openreview.net/forum?id=gZ9hCDWe6ke}
}

@inproceedings{li2022bevformer,
author = {Li, Zhiqi and Wang, Wenhai and Li, Hongyang and Xie, Enze and Sima, Chonghao and Lu, Tong and Qiao, Yu and Dai, Jifeng},
title = {BEVFormer: Learning Bird’s-Eye-View Representation from Multi-camera Images via Spatiotemporal Transformers},
year = {2022},
isbn = {978-3-031-20076-2},
publisher = {Springer-Verlag},
address = {Berlin, Heidelberg},
url = {https://doi.org/10.1007/978-3-031-20077-9_1},
doi = {10.1007/978-3-031-20077-9_1},
booktitle = {Computer Vision – ECCV 2022: 17th European Conference, Tel Aviv, Israel, October 23–27, 2022, Proceedings, Part IX},
pages = {1–18},
numpages = {18},
location = {Tel Aviv, Israel}
}

@inproceedings{yang2024unical_static_0,
  title={Unical: Unified neural sensor calibration},
  author={Yang, Ze and Chen, George and Zhang, Haowei and Ta, Kevin and B{\^a}rsan, Ioan Andrei and Murphy, Daniel and Manivasagam, Sivabalan and Urtasun, Raquel},
  booktitle={European Conference on Computer Vision},
  pages={327--345},
  year={2024},
  organization={Springer}
}

@misc{tahiraj2025calivlidartovehiclecalibrationarbitrary_static_1,
      title={CaLiV: LiDAR-to-Vehicle Calibration of Arbitrary Sensor Setups}, 
      author={Ilir Tahiraj and Markus Edinger and Dominik Kulmer and Markus Lienkamp},
      year={2025},
      eprint={2504.01987},
      archivePrefix={arXiv},
      primaryClass={cs.RO},
      url={https://arxiv.org/abs/2504.01987}, 
}

@ARTICLE{8057987_static_2,
  author={Gao, Yi and Lin, Chunyu and Zhao, Yao and Wang, Xin and Wei, Shikui and Huang, Qi},
  journal={IEEE Transactions on Intelligent Transportation Systems}, 
  title={3-D Surround View for Advanced Driver Assistance Systems}, 
  year={2018},
  volume={19},
  number={1},
  pages={320-328},
  doi={10.1109/TITS.2017.2750087}}

@INPROCEEDINGS{8814231_static_3,
  author={Baek, Iljoo and Kanda, Akshit and Tai, Tzu Chieh and Saxena, Anchan and Rajkumar, Ragunathan},
  booktitle={2019 IEEE Intelligent Vehicles Symposium (IV)}, 
  title={Thin-Plate Spline-based Adaptive 3D Surround View}, 
  year={2019},
  volume={},
  number={},
  pages={586-593},
  doi={10.1109/IVS.2019.8814231}}

@inproceedings{eker2022real_static_4,
  title={A Real-time 3D Surround View Pipeline for Embedded Devices.},
  author={Eker, Onur and Ercan, Burak and Bayraktar, Berkant and Bal, Murat},
  booktitle={VISIGRAPP (4: VISAPP)},
  pages={257--263},
  year={2022}
}

@inproceedings{dong2024dsvt_dynamic_0,
  title={DSVT: Dynamic 3D Surround View for Tractor-Trailer Vehicles Based on Real-Time Pose Estimation with Drop Model},
  author={Dong, Zhipeng and Fu, Mengyin and Liang, Hao and Zhu, Chunhui and Yang, Yi},
  booktitle={2024 IEEE/RSJ International Conference on Intelligent Robots and Systems (IROS)},
  pages={9461--9467},
  year={2024},
  organization={IEEE}
}

@INPROCEEDINGS{UDSV,
  author={Sun, Leyao and Liang, Hao and Dong, Zhipeng and Yang, Yi and Fu, Mengyin},
  booktitle={2025 IEEE International Conference on Robotics and Automation (ICRA)}, 
  title={UDSV: Unsupervised Deep Stitching for Tractor-Trailer Surround View}, 
  year={2025},
  volume={},
  number={},
  pages={5157-5163},
  doi={10.1109/ICRA55743.2025.11127831}}

@article{Unsupervised_deep_image_stitching,
title = {Unsupervised deep image stitching based on cascaded warping and multi-scale seam prediction for USV wide field-of-view generation},
journal = {Autonomous Transportation Research},
year = {2025},
issn = {3050-8622},
doi = {https://doi.org/10.1016/j.atres.2025.09.001},
url = {https://www.sciencedirect.com/science/article/pii/S3050862225000030},
author = {Zhilin Yang and Yong Yin and Qianfeng Jing and Zeyuan Shao and Haitong Xu and C. {Guedes Soares}}
}

@inproceedings{hu2023planning_uniad,
  title={Planning-oriented autonomous driving},
  author={Hu, Yihan and Yang, Jiazhi and Chen, Li and Li, Keyu and Sima, Chonghao and Zhu, Xizhou and Chai, Siqi and Du, Senyao and Lin, Tianwei and Wang, Wenhai and others},
  booktitle={Proceedings of the IEEE/CVF conference on computer vision and pattern recognition},
  pages={17853--17862},
  year={2023}
}

@inproceedings{jiang2023vad,
  title={Vad: Vectorized scene representation for efficient autonomous driving},
  author={Jiang, Bo and Chen, Shaoyu and Xu, Qing and Liao, Bencheng and Chen, Jiajie and Zhou, Helong and Zhang, Qian and Liu, Wenyu and Huang, Chang and Wang, Xinggang},
  booktitle={Proceedings of the IEEE/CVF International Conference on Computer Vision},
  pages={8340--8350},
  year={2023}
}

@inproceedings{liao2025diffusiondrive,
  title={Diffusiondrive: Truncated diffusion model for end-to-end autonomous driving},
  author={Liao, Bencheng and Chen, Shaoyu and Yin, Haoran and Jiang, Bo and Wang, Cheng and Yan, Sixu and Zhang, Xinbang and Li, Xiangyu and Zhang, Ying and Zhang, Qian and others},
  booktitle={Proceedings of the Computer Vision and Pattern Recognition Conference},
  pages={12037--12047},
  year={2025}
}

@InProceedings{song2025momad,
    author    = {Song, Ziying and Jia, Caiyan and Liu, Lin and Pan, Hongyu and Zhang, Yongchang and Wang, Junming and Zhang, Xingyu and Xu, Shaoqing and Yang, Lei and Luo, Yadan},
    title     = {Don't Shake the Wheel: Momentum-Aware Planning in End-to-End Autonomous Driving},
    booktitle = {Proceedings of the IEEE/CVF Conference on Computer Vision and Pattern Recognition (CVPR)},
    month     = {June},
    year      = {2025},
    pages     = {22432-22441}
}

@inproceedings{lu2025matrix3d,
  title={Matrix3d: Large photogrammetry model all-in-one},
  author={Lu, Yuanxun and Zhang, Jingyang and Fang, Tian and Nahmias, Jean-Daniel and Tsin, Yanghai and Quan, Long and Cao, Xun and Yao, Yao and Li, Shiwei},
  booktitle={Proceedings of the Computer Vision and Pattern Recognition Conference},
  pages={11250--11263},
  year={2025}
}

@InProceedings{Zhang_2025_CVPR_bridgeAD,
    author    = {Zhang, Bozhou and Song, Nan and Jin, Xin and Zhang, Li},
    title     = {Bridging Past and Future: End-to-End Autonomous Driving with Historical Prediction and Planning},
    booktitle = {Proceedings of the IEEE/CVF Conference on Computer Vision and Pattern Recognition (CVPR)},
    month     = {June},
    year      = {2025},
    pages     = {6854-6863}
}

@inproceedings{locher2016progressive_PMVS,
  title={Progressive prioritized multi-view stereo},
  author={Locher, Alex and Perdoch, Michal and Van Gool, Luc},
  booktitle={Proceedings of the IEEE conference on computer vision and pattern recognition},
  pages={3244--3252},
  year={2016}
}

@book{bundle,
author = {Hartley, Richard and Zisserman, Andrew},
title = {Multiple view geometry in computer vision},
year = {2000},
isbn = {0521623049},
publisher = {Cambridge University Press},
address = {USA}
}

@inproceedings{wang2024vggsfm,
  title={Vggsfm: Visual geometry grounded deep structure from motion},
  author={Wang, Jianyuan and Karaev, Nikita and Rupprecht, Christian and Novotny, David},
  booktitle={Proceedings of the IEEE/CVF conference on computer vision and pattern recognition},
  pages={21686--21697},
  year={2024}
}

@inproceedings{sun2021neuralrecon,
  title={Neuralrecon: Real-time coherent 3d reconstruction from monocular video},
  author={Sun, Jiaming and Xie, Yiming and Chen, Linghao and Zhou, Xiaowei and Bao, Hujun},
  booktitle={Proceedings of the IEEE/CVF conference on computer vision and pattern recognition},
  pages={15598--15607},
  year={2021}
}

@inproceedings{luu2025rc_AutoCalib,
  title={RC-AutoCalib: An End-to-End Radar-Camera Automatic Calibration Network},
  author={Luu, Van-Tin and Cai, Yon-Lin and Tran, Vu-Hoang and Chiu, Wei-Chen and Chen, Yi-Ting and Huang, Ching-Chun},
  booktitle={Proceedings of the Computer Vision and Pattern Recognition Conference},
  pages={6700--6709},
  year={2025}
}

@article{peebles2024dit,
  title={Scalable Diffusion Models with Transformers},
  author={William S. Peebles and Saining Xie},
  journal={2023 IEEE/CVF International Conference on Computer Vision (ICCV)},
  year={2022},
  pages={4172-4182},
  url={https://api.semanticscholar.org/CorpusID:254854389}
}

@inproceedings{herau2024soac,
  title={Soac: Spatio-temporal overlap-aware multi-sensor calibration using neural radiance fields},
  author={Herau, Quentin and Piasco, Nathan and Bennehar, Moussab and Roldao, Luis and Tsishkou, Dzmitry and Migniot, Cyrille and Vasseur, Pascal and Demonceaux, C{\'e}dric},
  booktitle={Proceedings of the IEEE/CVF Conference on Computer Vision and Pattern Recognition},
  pages={15131--15140},
  year={2024}
}

@inproceedings{cocheteux2024muli,
  title={MULI-EV: maintaining unperturbed lidar-event calibration},
  author={Cocheteux, Mathieu and Moreau, Julien and Davoine, Franck},
  booktitle={Proceedings of the IEEE/CVF Conference on Computer Vision and Pattern Recognition},
  pages={4579--4586},
  year={2024}
}

@inproceedings{10.1145/3664647.3680572_calibrbev,
author = {Liao, Wenlong and Qiang, Sunyuan and Li, Xianfei and Chen, Xiaolei and Wang, Haoyu and Liang, Yanyan and Yan, Junchi and He, Tao and Peng, Pai},
title = {CalibRBEV: Multi-Camera Calibration via Reversed Bird's-eye-view Representations for Autonomous Driving},
year = {2024},
isbn = {9798400706868},
publisher = {Association for Computing Machinery},
address = {New York, NY, USA},
url = {https://doi.org/10.1145/3664647.3680572},
doi = {10.1145/3664647.3680572},
booktitle = {Proceedings of the 32nd ACM International Conference on Multimedia},
pages = {9145–9154},
numpages = {10},
location = {Melbourne VIC, Australia},
series = {MM '24}
}

@inproceedings{zhou2025modeseq,
  title={ModeSeq: Taming Sparse Multimodal Motion Prediction with Sequential Mode Modeling},
  author={Zhou, Zikang and Zhou, Hengjian and Hu, Haibo and Wen, Zihao and Wang, Jianping and Li, Yung-Hui and Huang, Yu-Kai},
  booktitle={Proceedings of the Computer Vision and Pattern Recognition Conference},
  pages={1612--1621},
  year={2025}
}

@inproceedings{qin2024towards,
  title={Towards generalizable multi-object tracking},
  author={Qin, Zheng and Wang, Le and Zhou, Sanping and Fu, Panpan and Hua, Gang and Tang, Wei},
  booktitle={Proceedings of the IEEE/CVF Conference on Computer Vision and Pattern Recognition},
  pages={18995--19004},
  year={2024}
}
}

\clearpage
\setcounter{page}{1}
\maketitlesupplementary

\appendix

\noindent In this Appendix, we provide the following:
\begin{itemize}
    \item Implementation details in Appendix~\ref{sec: implement_details}.
    \item Additional dataset descriptions and visualizations in Appendix~\ref{sec:dataset}.
    \item Runtime and training parameters in Appendix~\ref{sec:runtime}.
    \item More qualitative results under different articulated driving scenarios in Appendix~\ref{sec:more_results}.
    \item Limitations and future work in Appendix~\ref{sec: limit}.
\end{itemize}

\section{Implementation Details}
\label{sec: implement_details}

\textbf{Metric-Scale Prediction.}
Dynamic trailer calibration requires camera poses expressed in metric units because the tractor–trailer linkage evolves in real distance and orientation. The magnitude of this inter-rig baseline determines the validity of multiview geometry, influences parallax and triangulation depth, and conditions the spatial accuracy of downstream modules such as BEV feature lifting, object localization, and motion estimation. Any ambiguity in scale introduces inconsistent baselines across frames, which directly degrades geometric alignment and accumulates as bias in downstream perception.

Unlike existing learning-based geometric methods, our framework dCAP directly predicts trailer rear camera poses in \textbf{metric scale} without requiring any post-hoc scale recovery. 
Current learning-based geometric models, including VGGT and DUSt3R, cannot provide such metric-scale poses. Their regression heads are optimized under a similarity-equivariant formulation due to limited availability and inconsistency of metric-supervised data across training sources. As a consequence, these models estimate camera motion only up to an arbitrary global scale. This yields relative geometry but omits absolute metric information, leading to per-sequence scale drift that prevents their direct use for articulated calibration, where the inter-rig translation must be recovered in meters at every frame.
To enable a fair comparison, we convert their normalized predictions into metric scale via a per-frame scale estimation procedure described below.

\textbf{Problem Setup.}
Let $T_i^\star \in \mathrm{SE}(3)$ denote the ground-truth extrinsic matrix of camera $i$ (tractor front, trailer rear, etc.) expressed in the world frame, and let $\hat{T}_i$ be the corresponding prediction from a baseline method. We use the tractor's front camera and the trailer's rear camera as a calibration pair. Their ground-truth relative transform is
\begin{equation}
T_{F \rightarrow B}^\star = (T_B^\star)^{-1} T_F^\star,
\end{equation}
and the baseline prediction yields
\begin{equation}
\hat{T}_{F \rightarrow B} = (\hat{T}_B)^{-1} \hat{T}_F.
\end{equation}
We decompose both into rotation and translation,
$T_{F \rightarrow B}^\star = (R^\star,\, t^\star)$ and
$\hat{T}_{F \rightarrow B} = (\hat{R},\, \hat{t})$,
and treat the unknown global scale as a scalar factor $s$ on the translation component.

\textbf{Scale Estimation.}
To recover a metric scale for the baseline prediction, we align the ground-truth and predicted relative translations by a one-dimensional least-squares fit. Concretely, we compute
\begin{equation}
s = \frac{(t^\star)^\top \hat{t}}{\hat{t}^\top \hat{t}},
\label{eq:scale_factor}
\end{equation}
which is the optimal scalar minimizing $\lVert t^\star - s \hat{t} \rVert_2^2$. The rotation $\hat{R}$ is left unchanged. This scale factor is recomputed independently for each frame, and we discard frames where $\|\hat{t}\|_2^2$ falls below a small threshold (indicating an unreliable baseline prediction).

\textbf{Metric-Scale Trailer Poses.}
Once the scale factor $s$ is obtained for a given frame, we apply it to all trailer-mounted cameras. For any trailer camera $j \in \{\text{rear}, \text{rear-left}, \text{rear-right}\}$, we first compute its baseline relative transform to the tractor front camera,
\begin{equation}
\hat{T}_{F \rightarrow j} = (\hat{T}_j)^{-1} \hat{T}_F = (\hat{R}_{F \rightarrow j},\, \hat{t}_{F \rightarrow j}),
\end{equation}
and then construct a metric-scale relative transform by rescaling only the translation:
\begin{equation}
\widetilde{T}_{F \rightarrow j} = (\hat{R}_{F \rightarrow j},\, s\, \hat{t}_{F \rightarrow j}).
\end{equation}
Finally, we map this metric-scale relative pose back to the world frame using the ground-truth tractor front extrinsic:
\begin{equation}
\widetilde{T}_j = T_F^\star \, \widetilde{T}_{F \rightarrow j}^{-1}.
\end{equation}
We extract the translation $\widetilde{t}_j$ and rotation $\widetilde{R}_j$ from $\widetilde{T}_j$ and use them as the metric-scale trailer camera extrinsics when evaluating VGGT and DUSt3R. The same procedure is applied to the trailer rear-left and rear-right cameras, using the known rigid intra-trailer transforms to maintain consistency.

\section{Dataset}
\label{sec:dataset}

\begin{figure}[t]
    \centering
    \includegraphics[width=1\linewidth]{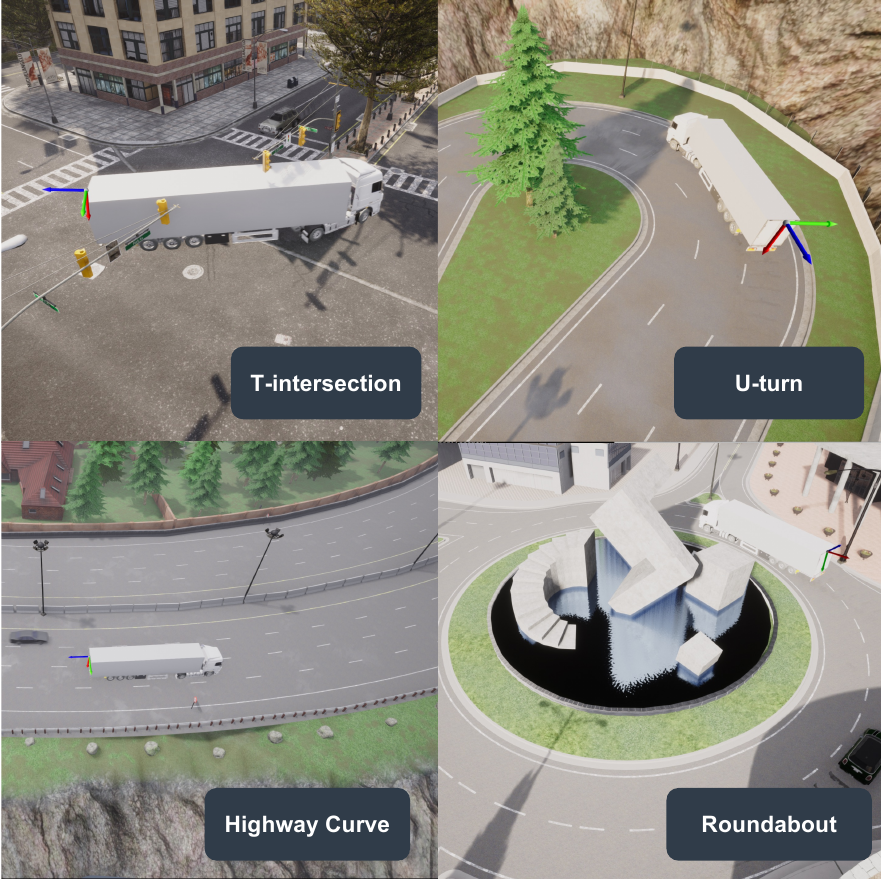}
    \caption{Representative articulated-driving scenarios in the STT4AT dataset. Each example shows the tractor–trailer configuration and trailer-mounted camera poses under typical high-articulation maneuvers.}
    \label{fig:differn_senarios}
\end{figure}

\begin{figure}
    \centering
    \includegraphics[width=1\linewidth]{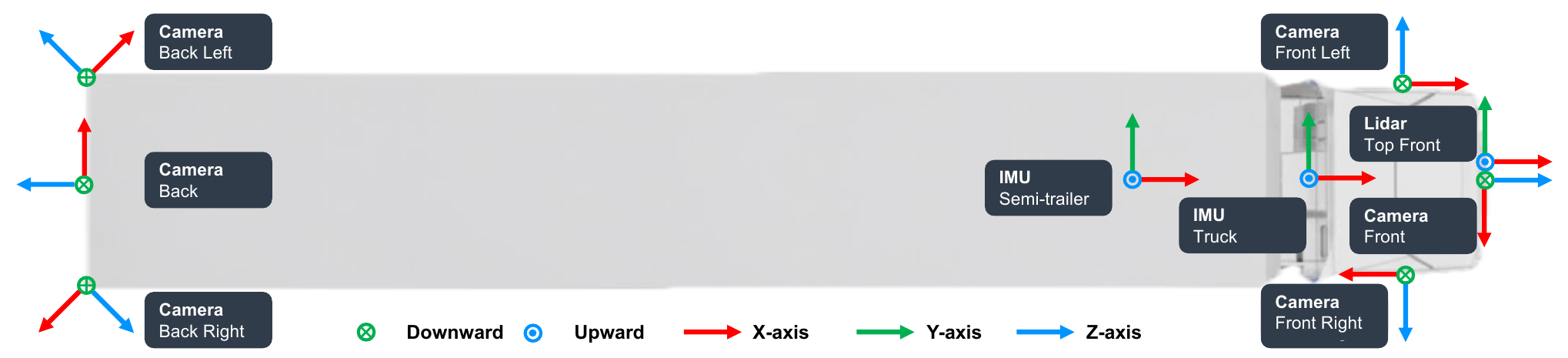}
    \caption{Placement of the sensors and their corresponding coordinate systems from a top-down view perspective}
    \label{fig:sensor_coord_system}
\end{figure}

\textbf{Sensor Setup.}
Figure~\ref{fig:sensor_coord_system} illustrates the placement of all onboard sensors.
The tractor is equipped with three forward-facing RGB cameras and a roof-mounted 128-beam LiDAR.
The trailer carries a rigid tri-camera rig mounted at its rear.
The extrinsics of tractor-mounted sensors remain fixed throughout each sequence, while the trailer-mounted rig undergoes time-varying motion due to articulation.

\textbf{Visualizations of Articulated Scenarios.}
Figure~\ref{fig:differn_senarios} shows representative articulated driving scenarios included in the dataset.
The collection spans common maneuvers such as T-intersections, U-turns, highway curves, and roundabouts.
These scenes cover a range of articulation magnitudes and occlusion patterns, providing diverse geometric conditions for evaluating calibration and perception models.

\section{Runtime and Memory}
\label{sec:runtime}

This section reports the computational footprint of the proposed model, including GPU memory usage, inference latency, and the number of learnable parameters.

\textbf{Training.}
Since the VGGT encoder remains frozen during training, only the CCA module, the CTA module, and the modulation–refinement head contribute to parameter updates. This design eliminates the need for heavy prediction branches such as depth estimation or dense point generation, resulting in a lightweight and computationally efficient framework. Table~\ref{tab:runtime_params} lists the number of trainable parameters for each component. 

During training, CTA requires the previous global token for temporal fusion. We randomly sample three consecutive frames; when no temporal neighbors exist (e.g., the first frame of a sequence), the model falls back to using the current frame’s own global token. 

\textbf{Inference.}
All experiments are performed on a single NVIDIA RTX~A6000. 
dCAP with CCA requires $8.5\,$GB of GPU memory during inference, while dCAP with CTA requires $10\,$GB. 
The inference time is $1.2\,$s per frame for both settings. 
CTA does not increase latency because temporal self-attention uses only a cached global token from the previous frame, and maintaining this cache lies outside the inference-time critical path.

\begin{table}[t]
\centering
\small
\setlength{\tabcolsep}{6pt}
\begin{tabular}{lrr}
\toprule
Module & \#Params (M) & Trainable \\
\midrule
VGGT backbone & 909.1 & \xmark \\
Camera Cross-Attention (CCA)                                                        & 16.8  & \cmark \\
Camera Temporal Self-Attention (CTA)                                                & 16.8  & \cmark \\
AdaLN-guided pose refinement module                                                  & 216.2 & \cmark \\
\bottomrule
\end{tabular}
\caption{Number of parameters for core modules in dCAP.}
\label{tab:runtime_params}
\end{table}

\section{More Results}
\label{sec:more_results}
\begin{figure}
    \centering
    \includegraphics[width=1\linewidth]{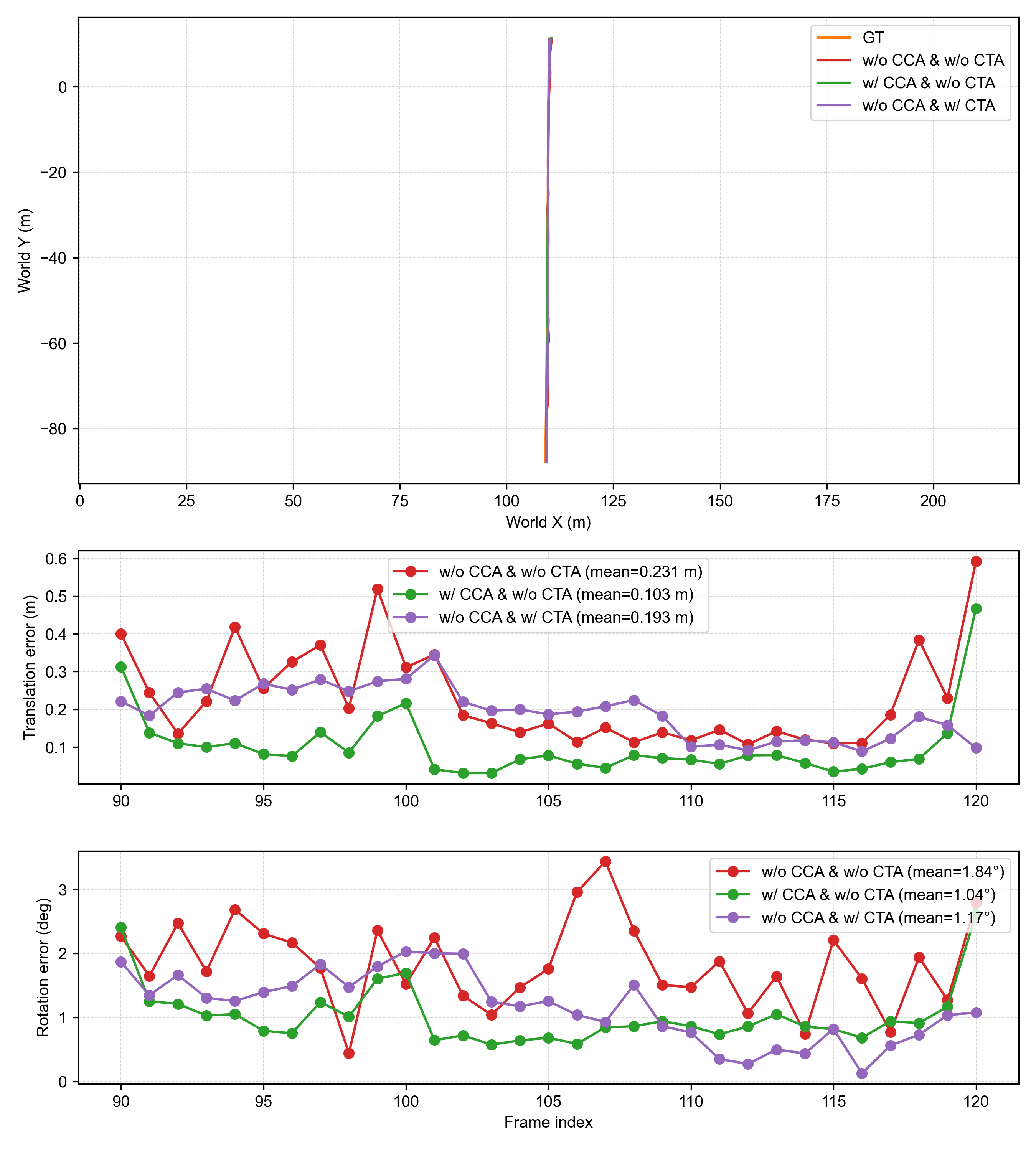}
    \caption{Qualitative results comparison between different attention modules under the straight scenario.}
    \label{fig:vis_straight}
\end{figure}

\begin{figure}
    \centering
    \includegraphics[width=1\linewidth]{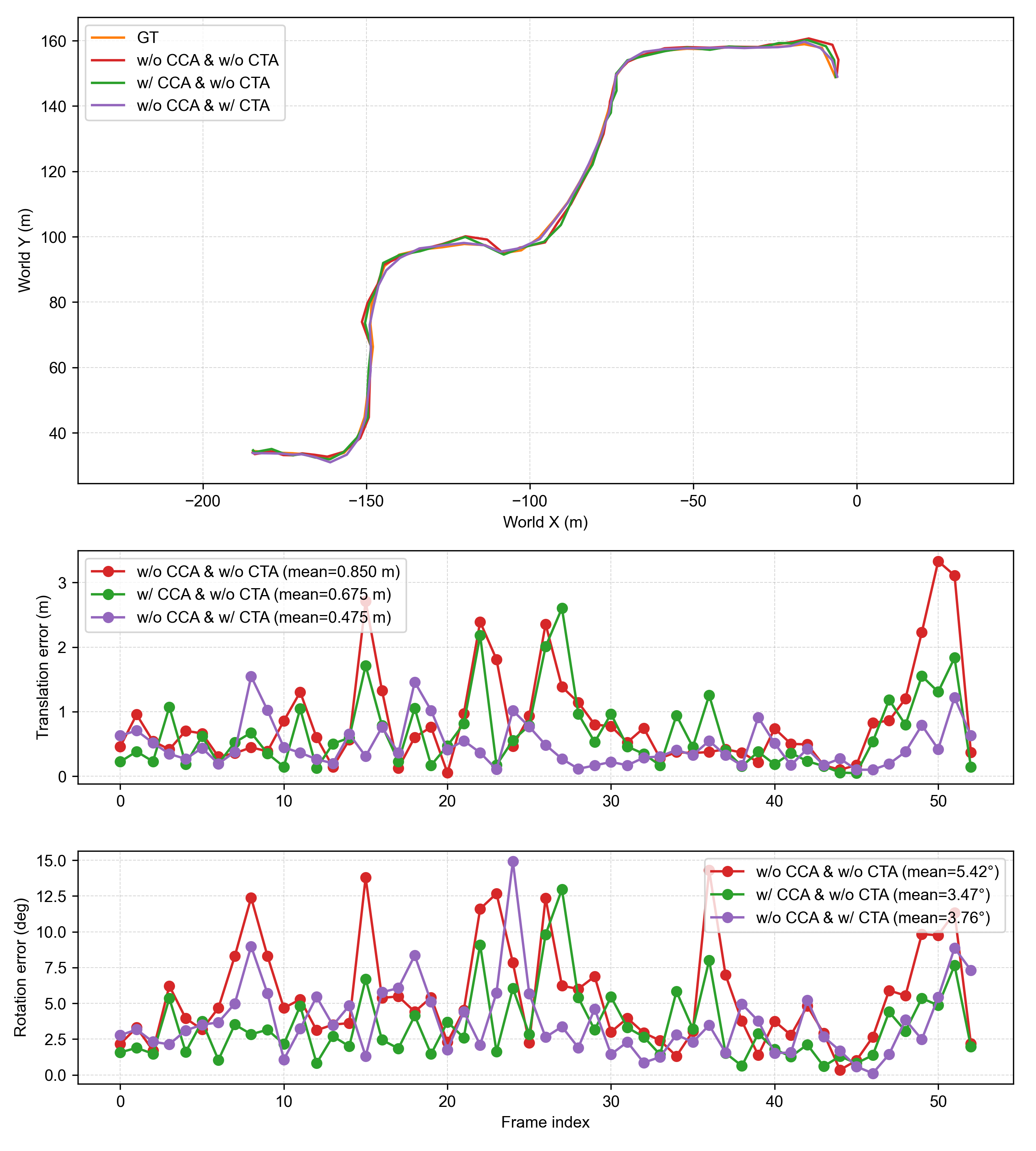}
    \caption{Qualitative results comparison between different attention modules under the roundabout scenario.}
    \label{fig:vis_roundabout}
\end{figure}

\begin{figure}
    \centering
    \includegraphics[width=1\linewidth]{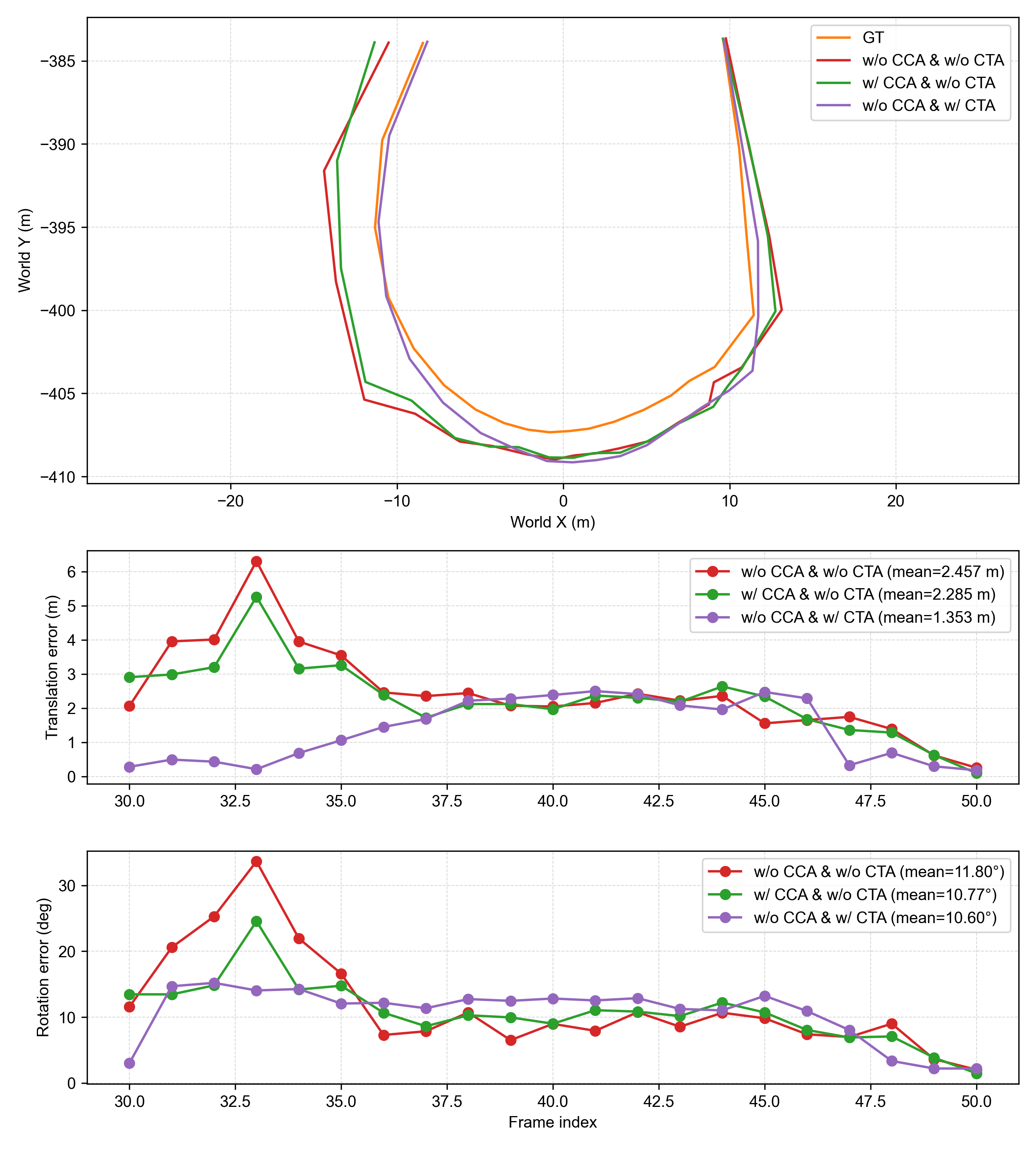}
    \caption{Qualitative results comparison between different attention modules under the u-turn scenario.}
    \label{fig:vis_u_turn}
\end{figure}

\begin{figure}
    \centering
    \includegraphics[width=1\linewidth]{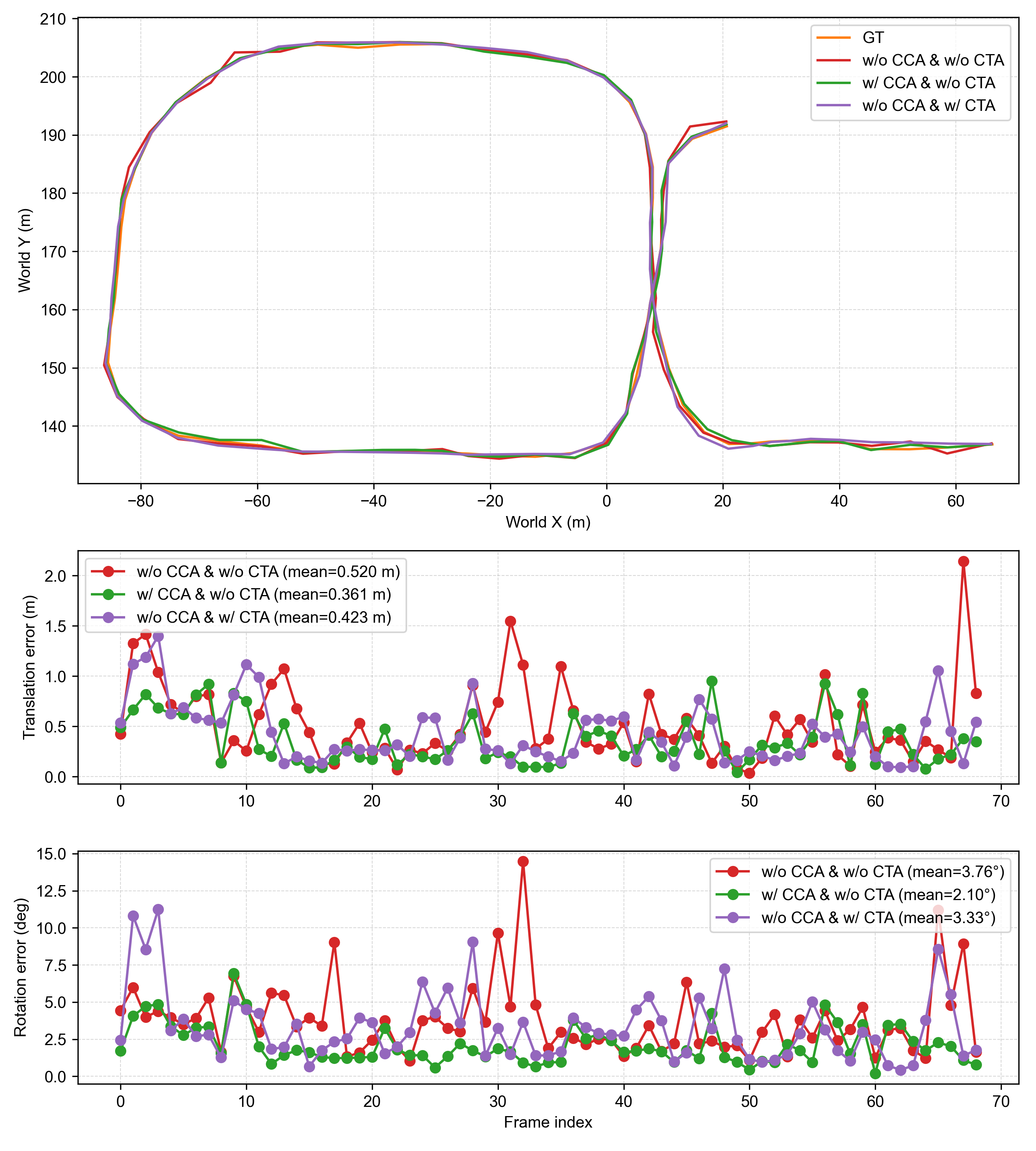}
    \caption{Qualitative results comparison between different attention modules under the multi-turn scenario.}
    \label{fig:vis_multi_turn}
\end{figure}

\textbf{More visual results.} In Figure~\ref{fig:vis_straight}, Figure~\ref{fig:vis_roundabout}, Figure~\ref{fig:vis_u_turn}, and Figure~\ref{fig:vis_multi_turn}, we demonstrate more visualization results of our method for different scenarios.

\section{Limitations and Future Work}
\label{sec: limit}
While STT4AT is equipped with synchronized multi-modal sensors supporting tasks from mapping to end-to-end driving, this paper  only focuses on establishing a baseline for 3D object detection. Moreover, the potential of the dataset for other downstream tasks (tracking, motion forecasting, mapping, and path planning) are still underexplored.

Furthermore, despite covering 87 scenes, the current dataset scale may limit generalization to long-tail articulated behaviors and rare environmental conditions. In real-world operations, a single tractor often couples with diverse trailer types, a complexity only partially captured here.

Future work will extend the benchmark to additional downstream tasks that are central to articulated navigation. We will also increase the variety of trailer configurations and scenarios to better support research on dynamic articulation.

\end{document}